\documentclass[letterpaper, 10 pt, conference]{ieeeconf}  

\IEEEoverridecommandlockouts                              

\usepackage{algorithm}
\usepackage{algorithmic}
\usepackage{xcolor}
\usepackage{url}
\usepackage{comment}
\usepackage{graphicx}
\usepackage{caption}
\usepackage{subcaption}
\usepackage{multirow}
\usepackage{svg}
\usepackage{adjustbox}
\usepackage{xcolor,colortbl}
\usepackage{amssymb}

\usepackage{xcolor,soul}

\title{\LARGE \bf 
Adaptive Human-Swarm Interaction based on Workload Measurement using Functional Near-Infrared Spectroscopy 
}

\author{Ayodeji O. Abioye$^{1}$*, Aleksandra Landowska$^{2}$*, William Hunt$^{1}$, Horia Maior$^{2}$, \\ Sarvapali D. Ramchurn$^{1}$, Mohammad Naiseh$^{3}$, Alec Banks$^{4}$, and Mohammad D. Soorati$^{1}$
\thanks{*These authors contributed equally.}
\thanks{$^{**}$Emails: \{a.o.abioye, w.hunt, sdr1, m.soorati\}@soton.ac.uk, \newline \{aleksandra.landowska, horia.maior\}@nottingham.ac.uk, \newline mnaiseh1@bournemouth.ac.uk, abanks@dstl.gov.uk}
\thanks{$^{1}$Electronics and Computer Science, University of Southampton, UK.}%
\thanks{$^{2}$School of Computer Science, University of Nottingham, UK.}%
\thanks{$^{3}$Computing and Informatics, Bournemouth University, UK}%
\thanks{$^{4}$Defence Science and Technology Laboratory, UK.}%
\thanks{$^{***}$The authors wish to acknowledge the support received from the EPSRC project on Smart Solutions Towards Cellular-Connected Unmanned Aerial Vehicles System (EP/W004364/1) and FAST-PI via the UKRI Trustworthy Autonomous Systems Hub (EP/V00784X/1).}%
}

\begin{document}

\maketitle
\thispagestyle{empty}
\pagestyle{empty}

\begin{abstract}

One of the challenges of human-swarm interaction (HSI) is how to manage the operator's workload. In order to do this, we propose a novel neurofeedback technique for the real-time measurement of workload using functional near-infrared spectroscopy (fNIRS). The objective is to develop a baseline for workload measurement in human-swarm interaction using fNIRS and to develop an interface that dynamically adapts to the operator's workload. The proposed method consists of using fNIRS device to measure brain activity, process this through a machine learning algorithm, and pass it on to the HSI interface. By dynamically adapting the HSI interface, the swarm operator's workload could be reduced and the performance improved. 
\end{abstract}

\section{Introduction}
The design of a suitable interface for human-swarm interaction for swarm operations can be challenging. The operator needs to be able to monitor the status of the swarm, which could consist of a thousand robots. The data collected by the swarm may need to be accessible to the operator in real-time without resulting in an information overload. One approach is to abstract the information to reduce workload. However, this could affect the explainability of the swarm actions and hence the operator's trust in the system.

Previous works investigating the effect of workload in human-swarm interaction (\cite{abioye2023,abioye_effect_2024,chandarana2021}) often rely on subjective user feedback using tools such as the NASA TLX survey questionnaire \cite{hart1986nasa}. This has the limitation that users are likely to provide feedback on elements that stood out during their experiments. Data on how the workload changes over time during the experiment is lost. When using self-reports, people reflect on past experiences that are not happening in real time. Moreover, a body of psychological research demonstrated that people are bad at expressing their mental states \cite{tempelaar2020subjective}. Therefore this research explores an alternative approach which uses functional near-infrared spectroscopy (fNIRS) to measure the swarm operator's workload over time more objectively. This information is then used to adapt the user interface to reduce the operator's workload.

\begin{figure}[!htb]
    \centering
    \frame{\includegraphics[width=0.9\columnwidth]{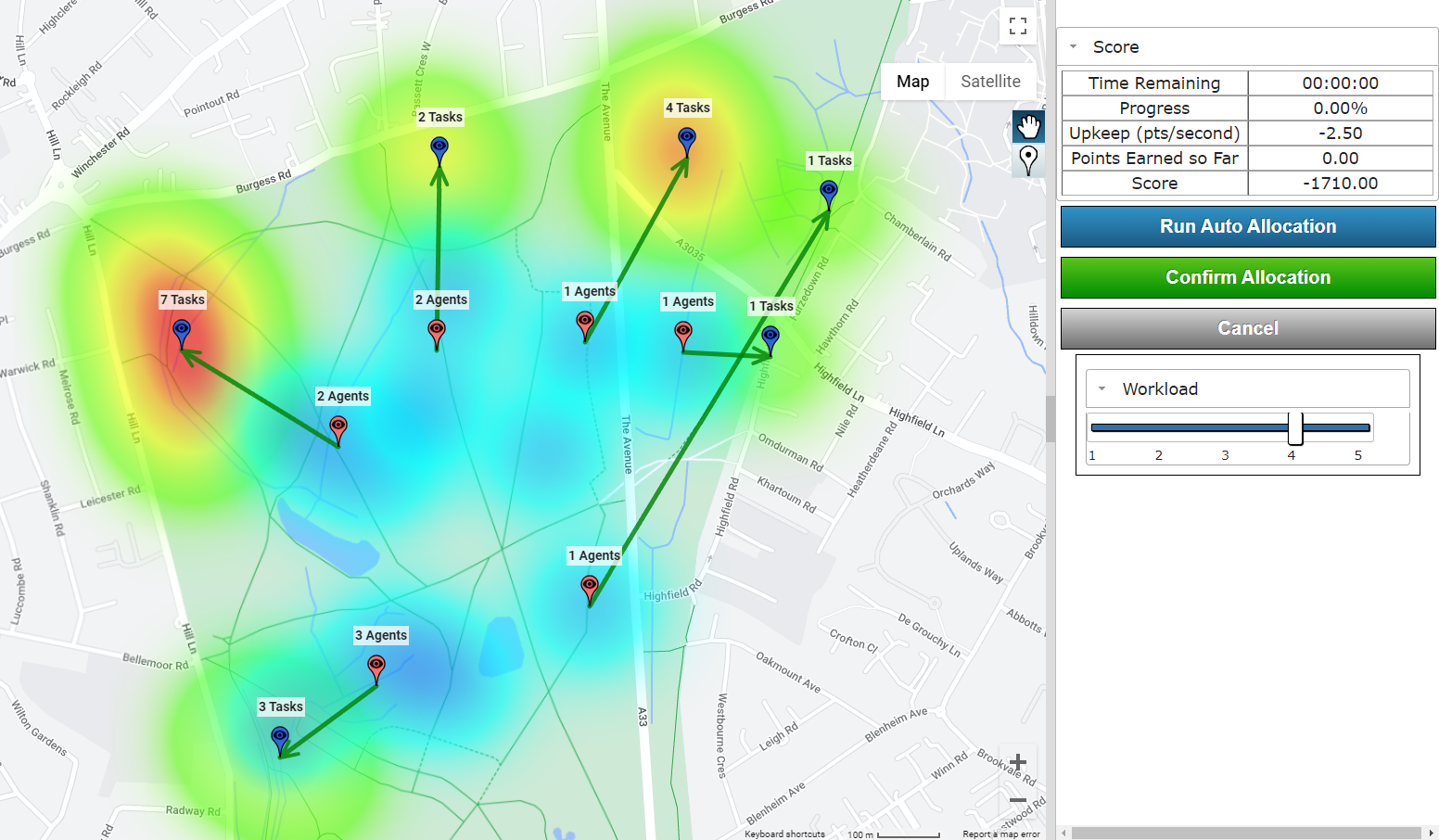}}
    \caption{Heatmap showing abstracted tasks and agent distribution to reduce operator workload.}
    \label{fig:adaptive_haris}
\end{figure}

\vspace{-0.5cm}

\section{Related Works}
Controlling swarms requires cognitive functions such as attention distribution, planning, working memory, decision-making, and mental workload. Understanding the mental state of the human operator through neurotechnology can increase safety and trust, and reduce errors in human-swarm operations. Real-time workload measurement has been done previously using pupillometry \cite{st2019planetary}. However, pupillometry (eye-tracking) for workload in swarms is unreliable, and the application is limited. This is because it is difficult to interpret, it is impacted by confounding variables such as the physiological state of the operator and lighting conditions, and it might obscure the view of the operator. 

Electroencephalography (EEG) was also used to measure mental workload indicators during the teleoperation of an unmanned aerial vehicle (UAV). The study demonstrated that EEG can identify different levels of mental workload in operators during human-swarm interaction scenarios \cite{Fernandez2020}.  However EEG suffers some drawbacks, such as poor spatial resolution, sensitivity to motion artifacts, and to environmental noise. In contrast, functional near-infrared spectroscopy (fNIRS) uses near-infrared light to measure brain activity, providing more robust signals that are less sensitive to motion and electromagnetic interference from the environment. fNIRS has previously been used to measure mental workload, demonstrating that it can effectively distinguish between different levels of workload, particularly in the prefrontal cortex, that are associated with variations in mental workload during realistic air traffic control and piloting tasks \cite{ayaz2012optical}, \cite{durantin2014using} or driving \cite{foy2018mental}. fNIRS has been used in measuring trust and mental workload in collaborative human-agent triads \cite{Eloy2022} and we examine its effectiveness in measuring workload in real-time during human-swarm interaction.

\section{Methodology}
The methodology consists of two parts. The first part focuses on developing the adaptive interface using the Human and Robot Interactive Swarm (HARIS) simulator \cite{SOORATI202493}. Heatmap representation was used to abstract agents and tasks using a five-point slider, similar to the trust slider used in \cite{nam2020}. Previous research \cite{divband2021designing} found this to be an effective representation when the swarm size or number of tasks is large. Figure \ref{fig:adaptive_haris} shows a screenshot of the developed human swarm interaction interface.

The second part focused on the application of the fNIRS in measuring workload during a HSI experiment in HARIS. We conducted a feasibility test on a single participant using the wireless fNIRS device. This was equipped with a probe for the frontal cortex consisting of eleven sources and seven detectors across twenty-three channels. The goal was to identify the brain area mediating HSI and key challenges or opportunities for using fNIRS in HSI. This sets the foundation for understanding the neural dynamics in HSI and highlights the potential of fNIRS in this field.

The user experiment consisted of two scenarios. In the first scenario, the participant matched individual drones to individual task. In the second scenario, heatmap was used to abstract the drones and tasks which meant that the participant could assign a cluster of agents to a cluster of tasks. The second scenario has more autonomy support that allows agents to intuitively distribute themselves to solving tasks even when the number of tasks outweighs the number of agents in the heatmap clusters. The aim was to determine which of the two interfaces imposes a greater mental workload on the user. The data collected in this study was analysed using a General Linear Model (GLM) approach, implemented through the NIRS Toolbox \cite{santosa2018nirs}, to assess the mental workload associated with each HARIS interface. 

\begin{figure}[!htb]
    \centering
    \frame{\includegraphics[width=0.9\columnwidth]{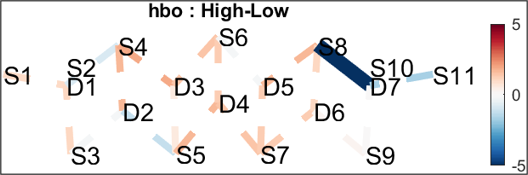}}
    \caption{Changes in HbO concentrations across prefrontal channels during interactions with the HARIS heatmap interface, as identified by fNIRS contrast analysis. The intensity of blue on a colour scale reflects the magnitude of these changes being statistically significant.}
    \label{fig:fnirs_workload_result}
\end{figure}

\vspace{-0.5cm}

\section{Result \& Discussion}
The results shown in Figure \ref{fig:fnirs_workload_result} revealed trends towards changes in the concentration of oxygenated haemoglobin (HbO) in the left dorsolateral prefrontal cortex (DLPFC), an area associated with mental workload, working memory, and decision-making. The increased activity in the left DPFC suggests that interacting with swarms, whether via a heat map or single-drone control, demands  cognitive resources from the user. This has implications for the design of user interfaces in swarm robotics, emphasizing the need to optimise these systems for minimal cognitive load while maintaining effective control and decision-making capabilities.

The application of neurotechnology in HSI also presents several challenges that need to be addressed. For instance, employing this technology in real-world scenarios introduces the difficulty of controlling confounding variables that might impact data quality. In real-world applications, the user's physical movements and changes in environmental lighting conditions could affect the measurements of brain activity, leading to inaccuracies in detecting mental workload levels. Addressing these challenges requires innovative solutions to ensure that neurotechnology can be effectively and reliably used outside of controlled laboratory settings.

Individual differences in brain responses are another significant challenge, as people have unique brain structures and exhibit unique brain signal patterns. This variability means that algorithms used to interpret these signals must be individually calibrated to each user, a process requiring extensive data collection that is often impractical in real-world settings due to time and resource constraints. To overcome these challenges in designing better autonomous systems, employing AI could be a promising approach. AI and machine learning techniques could adapt to individual differences more efficiently by learning from smaller datasets. Such approaches could enable systems to quickly adjust to new users, reducing the need for extensive calibration while still providing personalized and effective solutions for HSI.

Acquiring ethics approval to work on the sensitive brain signals of the users is a complex process due to the privacy concerns of accessing and processing personal data. Due to this limitation, the current work only contains information from one case study of one of the authors, hence it is not generalisable. The purpose of the current work was to demonstrate the methodology and test that it is possible to measure detectable effects. Future works would explore whether this detection is generalisable by recruiting more participants.

\section{Conclusions}
In this paper, we proposed a method to improve the human-swarm interaction experience for swarm operators by reducing their workload while maintaining high-performance levels. We defined a human swarm teaming task requiring varying levels of focus and dexterity resulting in high and low stress/workload conditions. We tested the task feasibility for detecting workload using fNIRS. We highlighted the challenges in deploying fNIRS for measuring workload. Further works would (a) conduct studies to develop the baseline for fNIRS workload measurement in HSI, and (b) integrate workload measurement directly into the HSI interface, so it can dynamically adapt to changes in workload due to task density increasing or decreasing while maintaining a high task completion efficiency.

\addtolength{\textheight}{-13cm}   







\bibliographystyle{IEEEtran}
\bibliography{ref}

\end{document}